\documentclass[conference]{IEEEtran}
\IEEEoverridecommandlockouts
\usepackage{cite}
\usepackage{amsmath,amssymb,amsfonts}
\usepackage{algorithmic}
\usepackage{graphicx}
\usepackage{textcomp}
\usepackage{subfigure}
\usepackage{xcolor}
\def\BibTeX{{\rm B\kern-.05em{\sc i\kern-.025em b}\kern-.08em
    T\kern-.1667em\lower.7ex\hbox{E}\kern-.125emX}}
\begin{document}

\title{ Behavioral Differences is the Key of Ad-hoc Team Cooperation in Multiplayer Games Hanabi


}

\author{\IEEEauthorblockN{Hyeonchang Jeon}
\IEEEauthorblockA{\textit{Artificial Intelligence Graduate School} \\
\textit{Gwangju Institute of Science and Technology (GIST)}\\
Gwangju, Korea \\
kevinjeon119@gm.gist.ac.kr}
\and
\IEEEauthorblockN{* Kyung-Joong Kim}
\IEEEauthorblockA{\textit{School of integrated technology} \\
\textit{Gwangju Institute of Science and Technology (GIST)}\\
Gwangju, Korea \\
kjkim@gist.ac.kr}
}

\maketitle

\begin{abstract}
Ad-hoc team cooperation is the problem of cooperating with other players that have not been seen in the learning process. Recently, this problem has been considered in the context of Hanabi, which requires cooperation without explicit communication with the other players. While in self-play strategies cooperating on reinforcement learning (RL) process have shown success, there is the problem of failing to cooperate with other unseen agents after the initial learning is completed. In this paper, we categorize the results of ad-hoc team cooperation into $Failure$, $Success$, and $Synergy$ and analyze the associated failures. First, we confirm that agents learning via RL converge to one strategy each, but not necessarily the same strategy, and that these agents can deploy different strategies even though they utilize the same hyperparameters. Second, we confirm that the larger the behavioral difference, the more pronounced the failure of ad-hoc team cooperation, as demonstrated using hierarchical clustering and Pearson correlation. We confirm that such agents are grouped into distinctly different groups through hierarchical clustering, such that the correlation between behavioral differences and ad-hoc team performance is -0.978. Our results improve understanding of key factors to form successful ad-hoc team cooperation in multi-player games.
\end{abstract}

\begin{IEEEkeywords}
Hanabi, Ad-hoc cooperation, Multi-agent Reinforcement Learning
\end{IEEEkeywords}

\section{Introduction}

Many researches in Reinforcement Learning (RL) have successfully solved various problems in games such as Go\cite{silver2016mastering} and Starcraft II\cite{vinyals2019grandmaster}. The board game Hanabi, seen in Figure 1, provides new challenges in the multi-agent domain, given its distinctive features from other multi-agent environments. Although Hanabi is a cooperative game, it inhibits communication between players, forcing the game to interpret the implicit meaning of other players' actions. To understand this implicit meaning of other players, modeling of other players is used.

\begin{figure}[ht]
\centerline{\includegraphics[width=0.49\textwidth]{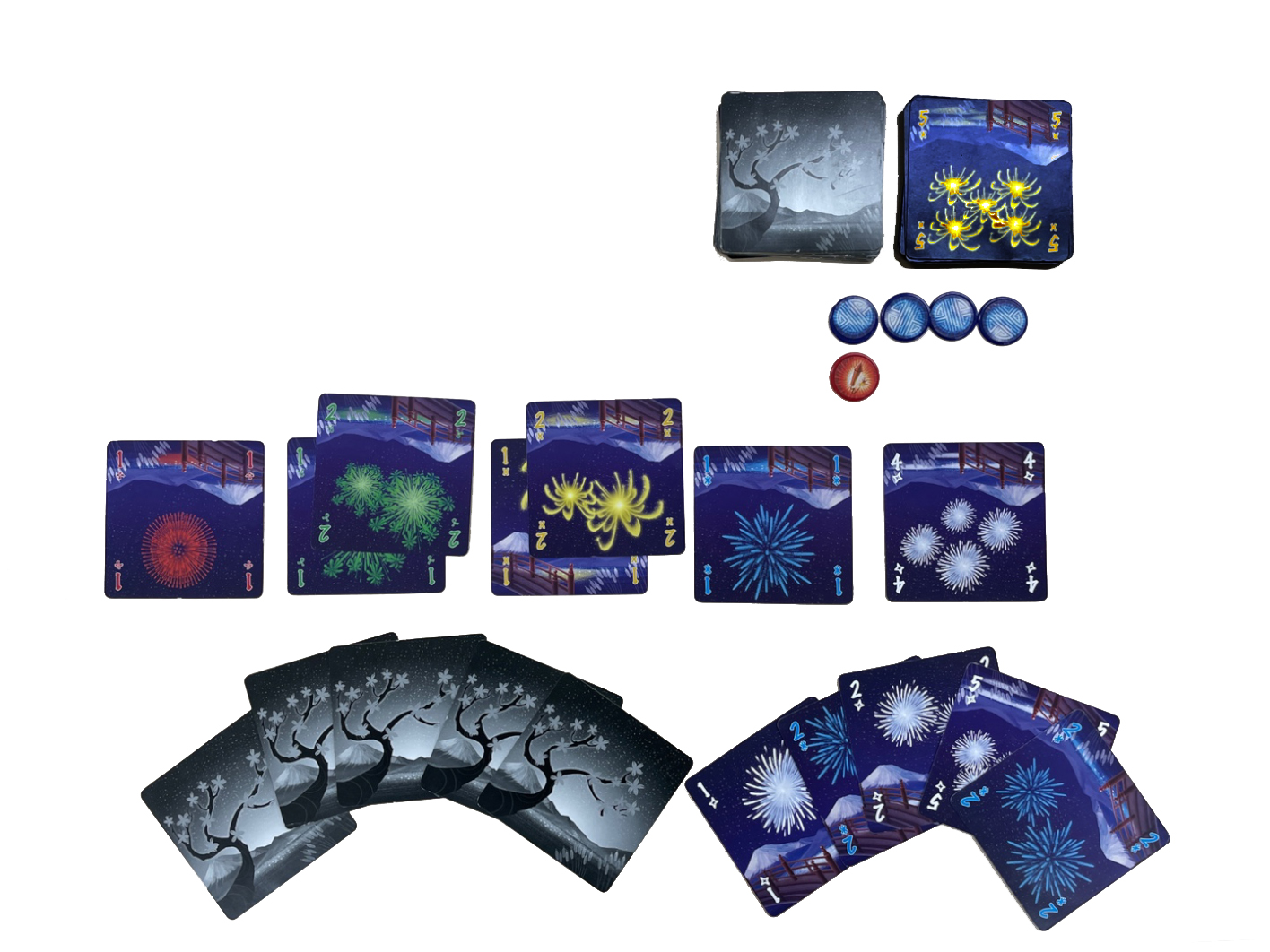}}
\label{fig1}
\caption{Photo of the card game Hanabi}
\end{figure}

The Theory of Mind (ToM)\cite{rabinowitz2018machine} is used to model the others' behaviors, beliefs, and intentions. By using the ToM, Bayesian Action Decoder (BAD) \cite{foerster2019bayesian} successfully played Hanabi with two players in a self-play setting. However, there are limitations when determining a player's beliefs, particularly when an agent trained in self-play settings cooperates with unseen agents, called ad-hoc team cooperation. Since an agent trained via self-play scenarios has no prior knowledge regarding the unseen agents, this trained agent fails to play effectively with the unseen agents.

There has been discussion on ad-hoc team play of agents trained by RL, especially in self-play settings, \cite{bard2020hanabi}\cite{canaan2020evaluating} but there is still limited understanding regarding the reasons for success and failure of cooperation in ad-hoc team play. For example, \cite{bard2020hanabi} reported that Actor Critic Hanabi Agent (ACHA), based on the Importance Weighted Actor Learner architecture (IMPALA)\cite{espeholt2018impala}, failed to cooperate with unseen agents despite ACHA successfully maneuvering in self-play settings. On the other hand, Rainbow\cite{hessel2018rainbow}, which is a value-based RL agent, played well under ad-hoc cooperation with other Rainbow agents (trained with different random seeds). Although the Rainbow agent was successful with other Rainbow agents, it was found not successful in ad-hoc team play with unseen rule-based agents\cite{canaan2020evaluating}.

\begin{table*}[t]
\renewcommand{\arraystretch}{1.3}
\caption{Summary of Related Works}
\label{jeon.t1}
\centering
\setlength{\tabcolsep}{3pt}
\begin{tabular}{|c|c|c|c|}
\hline
 Authors & Environment & Problem & Method \\ 
\hline
 Walton $et \ al.$\cite{walton20192018}, O’Dwyer $et \ al.$\cite{o2020github}, Cox $et \ al.$\cite{cox2015make},Wu $et \ al.$\cite{wu2018state} & Hanabi & Self-Play & Rule-based \\ 
 Hu $et \ al.$\cite{hu2019simplified}, Lerer $et \ al.$\cite{lerer2020improving} & Hanabi & Self-Play & Reinforcement Learning \\
 Cannan $et \ al.$\cite{canaan2019diverse} & Hanabi & Ad-hoc Play & Evolutionary methods \\
 Barrett $et \ al.$\cite{barrett2011empirical} & Pursuit & Ad-hoc Play & Planning method\\
 Agmon $et \ al.$\cite{agmon2014modeling} & Matrix Game & Ad-hoc Play & REACT algorithm\\
 Barrett $et \ al.$\cite{barrett2017making} & Half Field Offense, Pursuit & Ad-hoc Play & Reinforcement Learning\\
 Chen $et \ al.$\cite{chen2020aateam} & Half Field Offense & Ad-hoc Play & Reinforcement Learning\\
 Ravula $et \ al.$\cite{ravula2019ad} & Pursuit & Ad-hoc Play & MCTS \\

 \hline
\end{tabular}
\end{table*}

In this paper, we propose a method of analyzing the ad-hoc cooperation of agents trained by RL. This helps in understanding the reasons behind successful ad-hoc cooperation of agents. We found that the Rainbow agents often resulted in very similar players even with different random seeds. This makes ad-hoc team plays between Rainbow agents quite successful when working together, but is not very effective with unseen rule-based agents. In more detail, we categorize the ad-hoc cooperation results into three groups, $Failure$, $Success$, and $Synergy$. These are defined by comparing the self-play and ad-hoc performances. For failures, we applied hierarchical clustering and Pearson correlation analysis revealing that behavioral differences are key to successful ad-hoc cooperation.

\section{Backgrounds}
\label{relatedWorks}

This section introduces ad-hoc cooperation and Hanabi cooperation. Table \ref{jeon.t1} summarizes related works.

\subsection{The Cooperative Card Game Hanabi} 
In this section, we introduce the overall gameplay mechanics of Hanabi and its associated research values. Hanabi is a cooperative, partially observable, multiplayer game with 2 to 5 players. The game consists of a total of 25 five colors($R$, $Y$, $G$, $B$, $W$) and five ranks($1$, $2$, $3$, $4$, $5$), with numbers distributed differently depending on the rank. There are two $2$s, $3$s, and $4$s, one $5$ and three $1$s, for a total of 50 cards in deck. There are eight information tokens and three life tokens that are managed by the players. Players obtain higher scores by stacking the same color cards in order of rank.

During the game, the players can take different actions as follows: $Play$, $Hint Color$, $Hint Rank$, and $Discard$. The action $Play$ lets the player select a card from their hand, followed by determining whether the card is a score or a fail. For a failure, the player loses one life token. If not, the card is stacked, increasing the score. The action $Discard$ allows a player to discard a selected card from their hand, then recover an information token which is used to provide hints to the others, and then the player draws a new card from the deck. The action $Hint Color$ allows a player to use one hint token, allowing them to give another player a hint regarding a particular color. The player who receives the hint is given two positive information statements about which cards are which colors, as well as which colors do not correspond to a given card. Likewise, the action $Hint Rank$ also the player to use one hint token to provide another player similar information as for $Hint Color$, but instead regarding information on the rank rather than color.

Lastly, the game ends in three ways. First, when the players obtain a total score of 25. Second, when there are no remaining cards in the deck, the players take their last turn to play cards and increase the score of the stacked cards. Third, when the players fail to play a total of three times, the game ends and a score of 0 obtained. Therefore, the score ranges from 0 to 25.

Hanabi has the following characteristics relevant to research. First, Hanabi is a cooperative game, but unlike other cooperative games, communication between players is prohibited. Therefore, players must interact with other players implicitly through actions. Here, the players not only need to communicate their intentions correctly through their actions, but must also understand the intentions of the other players. Secondly, this game is partially observable. However, unlike other partially observable games that do not allow game information of other players, players in Hanabi instead do not know their own information. Since this information directly connects to players' actions, cooperation in Hanabi depends on the other players in an intricate way.

\subsection{Hanabi Cooperation}
There are several studies on Hanabi. In the case of rule-based agents, Hanabi competition\cite{walton20192018} introduces several rule-based agents participated in Hanabi competition. For such competition, the agents were implemented methods such as rule-based, Monte Carlo Tree Search(MCTS), and evolutionary methods such as genetic algorithms. HLE (Hanabi Learning Environment) \cite{bard2020hanabi} used SmartBot\cite{o2020github}, HatBot\cite{cox2015make}, and WTFWThat\cite{wu2018state} to compare the rule-based agents with different RL algorithms. Here, HLE implemented ACHA and Rainbow to compare their performance in self-play settings. BAD used the public belief Markov Decision Process(PuB-MDP) to construct a public belief and thus construct the agent models. Simplified Action Decoder(SAD)\cite{hu2019simplified} achieved an average score of 24.08 out of 25, which promoted more exploratory actions. Search for Partially Observing Teams of Agents(SPARTA)\cite{lerer2020improving} used a multi-agent policy search and achieved a state-of-the-art average score of 24.61.

Another problem in Hanabi is ad-hoc team cooperation which is found in the HLE. In HLE, the ACHA agents who were trained under self-play setting, failed to cooperate with other ACHA agents who were trained with different machines. In the case of Rainbow agents, they were able to obtain decent scores in HLE. However, in \cite{canaan2020evaluating}, the failure of ad-hoc cooperation between Rainbow and rule-based agents was similar to our initial findings, where we found that Rainbow agents failed to cooperate with agents that implement different strategies. \cite{canaan2019diverse}uses MAP-Elites to generate diverse agents for ad-hoc team settings.

\subsection{Ad-hoc Team Cooperation}
Ad-hoc team cooperation refers to an agent team of autonomous agents performing a task without any predefined cooperation \cite{stone2010ad}. Traditionally, Ad-hoc team cooperation was used in simple domains such as pursuit\cite{barrett2011empirical} and utility matrices\cite{agmon2014modeling}. These domains are fully observable and have low-dimensions and action spaces. Further, traditional approaches predefined the behavioral type of the agents so that they could not cooperate with unseen behavioral types among different agents. Recently, there have been many studies on more complex domains such as Half Field Offense(HFO)\cite{barrett2017making} based on the RoboCup 2D domain\cite{hausknecht2016half}, but these studies focused on past teammates with predefined types. There are also studies regarding ad-hoc teamwork with interesting methods such as the attention method in \cite{chen2020aateam}, change point detection(CPD) in \cite{ravula2019ad}.

However, the main difference between existing ad-hoc cooperation domains and Hanabi is the ambiguity of agent roles. Since roles are not explicitly defined in the game, it may seem that everyone plays the same role, but this is not the case. This is because the most important feature of Hanabi is coordinating agents, which depends strongly on the other agents. Thus, in order for the game to end successfully, roles must be divided evenly. In particular, if an agent takes certain actions excessively, the other agents are forced to play accordingly. For example, if one agent utilizes too many $Play$ actions, the other agent must focus on providing hints to prevent the game from ending, or alternatively use the $Discard$ action to discard a card to properly manage hint tokens. These roles are not predetermined by the agents, but rather by the agents while playing. As such, these roles are not fixed, and agents may have to change roles periodically depending on the situation. Furthermore, there are many cases in Hanabi that are further limited by additional rules. This key difference makes Hanabi distinct from other ad-hoc cooperation domains.

\begin{table*}[t]
\renewcommand{\arraystretch}{1.3}
\caption{Composition of Rule-based agents}
\label{jeon.t2}
\centering
\begin{tabular}{|c|c|c|c|c|c|c|}
\hline
 Rules/Agents & Internal Agent & Outer Agent & Van den Bergh Agent & Piers Agent & Flawed Agent & IGGI Agent\\ [0.1ex] 
\hline
 PlaySafeCard & O & O &  & O & O & O \\
\hline
 OsawaDiscard  & O & O &  & O & O & O \\
\hline
 TellPlayableCard  & O &  &  &  &  &  \\
\hline
 TellDispensable  &  &  &  & O &  &  \\
\hline
 PlayProbablySafeCard(0.25)  &  &  &  &  & O &  \\
\hline
 DiscardOldestFirst  &  &  &  & O & O & O \\
\hline
 TellPlayableCardOuter  &  & O &  &  &  &  \\
\hline
 TellUnknown &  & O &  &  &  &  \\
\hline
 PlayIfCertain &  &  &  &  &  & O \\
\hline
 HailMary &  &  &  & O &  &  \\
\hline
 PlayProbablySafeCard(0.6) &  &  & O & O &  &  \\
\hline
 TellAnyoneAboutUsefulCard &  &  & O &  &  & O \\
\hline
 DiscardProbablyUselessCard &  &  & O &  &  &  \\
\hline
 TellAnyoneAboutUselessCard &  &  & O &  &  &  \\
\hline
 TellMostInformation &  &  & O &  &  &  \\
\hline
 DiscardProbablyUselessCard(0.0) &  &  & O &  &  &  \\
\hline
 TellRandomly & O &  &  & O & O &  \\
\hline
 DiscardRandomly & O & O &  & O & O &  \\
\hline
\end{tabular}
\end{table*}

\section{Ad-hoc Team Play Analysis}
\label{Ad-hoc Team Play Analysis}

\subsection{Categorization of Ad-hoc Team Play}
Before analyzing ad-hoc team cooperation, we define the results of ad-hoc team cooperation according to three outcomes: $Failure$, $Success$, and $Synergy$. We define the function $AdHoc$ as the ad-hoc cooperation performance of agent $a_1$ and $a_2$ for $a_1, a_2 \in A$. The function $SelfPlay$ represents the self-play performance of agent $a \in A$. For example, let agent $a_1$ has a self-play score of 20.1 and agent $a_2$ a self-play score of 20.8. If $a_1$ and $a_2$ obtain an ad-hoc play score of 17.3, this ad-hoc performance is classified as a $Failure$ because the agents obtained a worse score than their individual self-play scores. If $a_1$ and $a_2$ instead obtain an ad-hoc play score of 20.5, the result is considered a $Success$ since one of the agents obtained a better score, even if the other did not. Lastly, if an ad-hoc play score of 23.2 is obtained, the result is classified as $Synergy$ since both agents outperform the individual self-play scores. The $Failure$ condition is following: 

\small
\begin{equation}
 AdHoc(a_1, a_2) < min(SelfPlay(a_1),SelfPlay(a_2))\label{eq1}
\end{equation}
\normalsize 

The $Success$ of ad-hoc play lies between the minimum and maximum values of the self-play scores of each two agents, defined by concepts such as teacher and learner. To simplify, we define $MinSelfPlay$ and $MaxSelfPlay$ from \ref{eq1} as follows.

\small 
\begin{equation}
MinSelfPlay = min(SelfPlay(a_1),SelfPlay(a_2)) \label{eq2}
\end{equation}

\begin{equation}
MaxSelfPlay = max(SelfPlay(a_1),SelfPlay(a_2)) \label{eq3}
\end{equation}

\begin{equation}
MinSelfPlay \leq AdHoc(a_1, a_2) \leq MaxSelfPlay \label{eq4}
\end{equation}
\normalsize

Lastly, the $Synergy$ condition occurs when two agents perform better than the maximum self-play performances.

\small
\begin{equation}
MaxSelfPlay < AdHoc(a_1, a_2) \label{eq5}
\end{equation}
\normalsize

\subsection{Analysis by Behavioral Difference}

To analyze the ad-hoc performance, we use three analyzing methods, hierarchical clustering, Pearson correlation, in conjunction with behavioral differences. Hierarchical clustering groups individuals based on similarities. This is especially useful when it is unknown how many groups will be determined. We cluster these agents to show the differences between Rainbow agents only and a mixture of Rainbow agents and rule-based agents. To analyze the ad-hoc team cooperation failure of Rainbow agents, we analyzed the differences in optimal behaviors as judged by each agent for the observations that came out through sampling. First, we replace the actions with one-hot vectors and then calculate the behavioral differences. We define the Behavioral Difference(BD) as follows:

\begin{equation}
BD = \sum_{n\in N} abs(x^n-y^n)\ for\ \forall x^i,y^i \in X,Y\label{eq6}
\end{equation}

For example, let agent $X$ and agent $Y$ have the different actions for three states and the same actions for two states. We obtain a BD of 6 since there are 2 measures for each different state. Finally, we use Pearson correlation coefficients to explain the behavioral differences and ad-hoc cooperation performance based on the behavioral differences as judged by Rainbow agents on the same state, which a significant impact on ad-hoc team cooperation.

\begin{figure*}
\centerline{\includegraphics[width=0.95\textwidth]{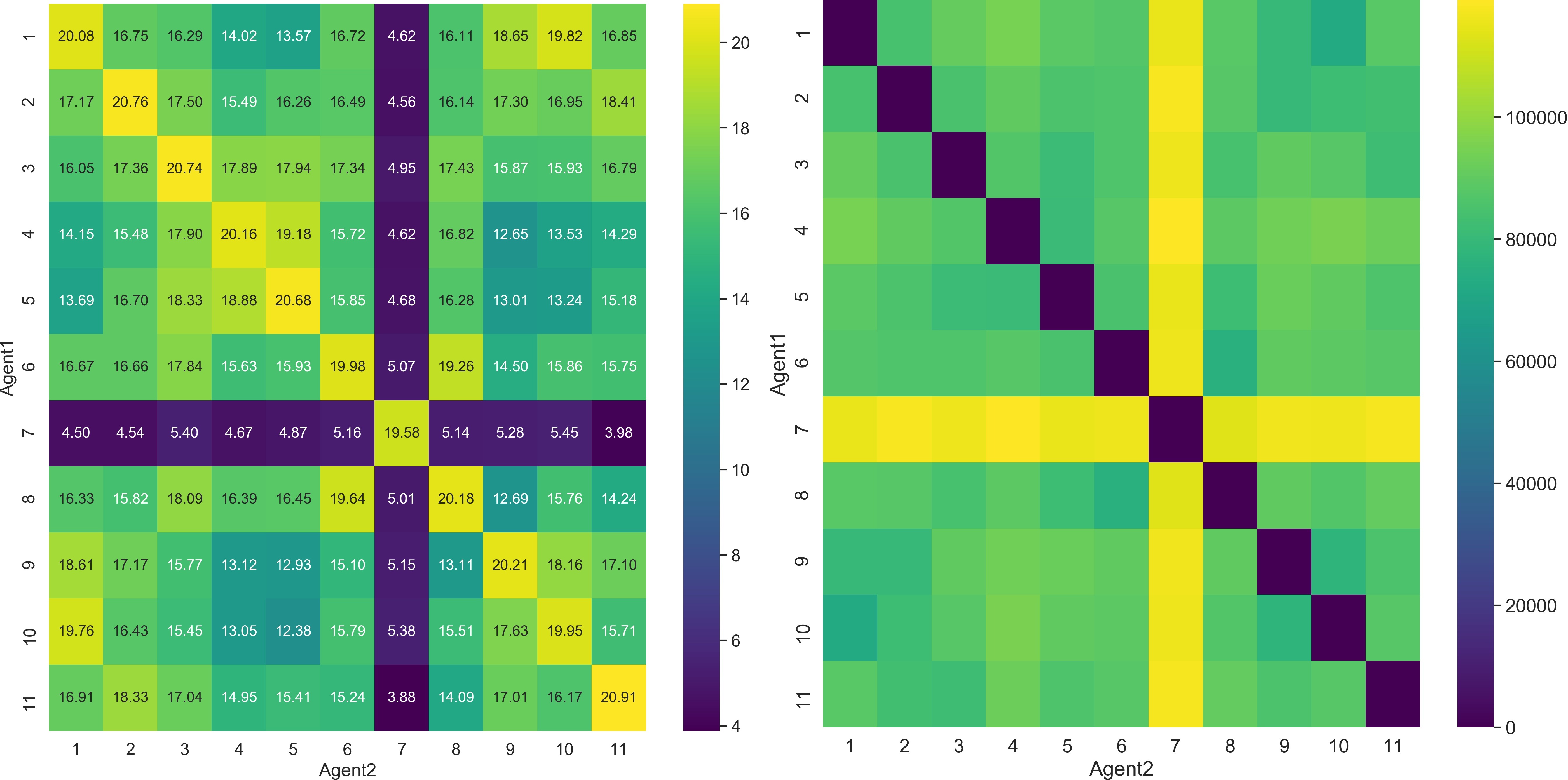}}
\label{fig2}
\caption{The left figure shows the average score of a two-player Hanabi game for 1,000 games. Each diagonal element represents the self-play score of each agent, which was almost 20. Since Hanabi is a turn-based game, when playing ad-hoc team games, the agent is divided into the order of the agents and an ad-hoc cooperative experiment is conducted. The other elements refer to the ad-hoc cooperation score. The right figure means the action decision differences of each agent for approximately 86,000 states of the sampled 1,100 games.}
\end{figure*}

\begin{figure*}
\centerline{\includegraphics[width=0.95\textwidth]{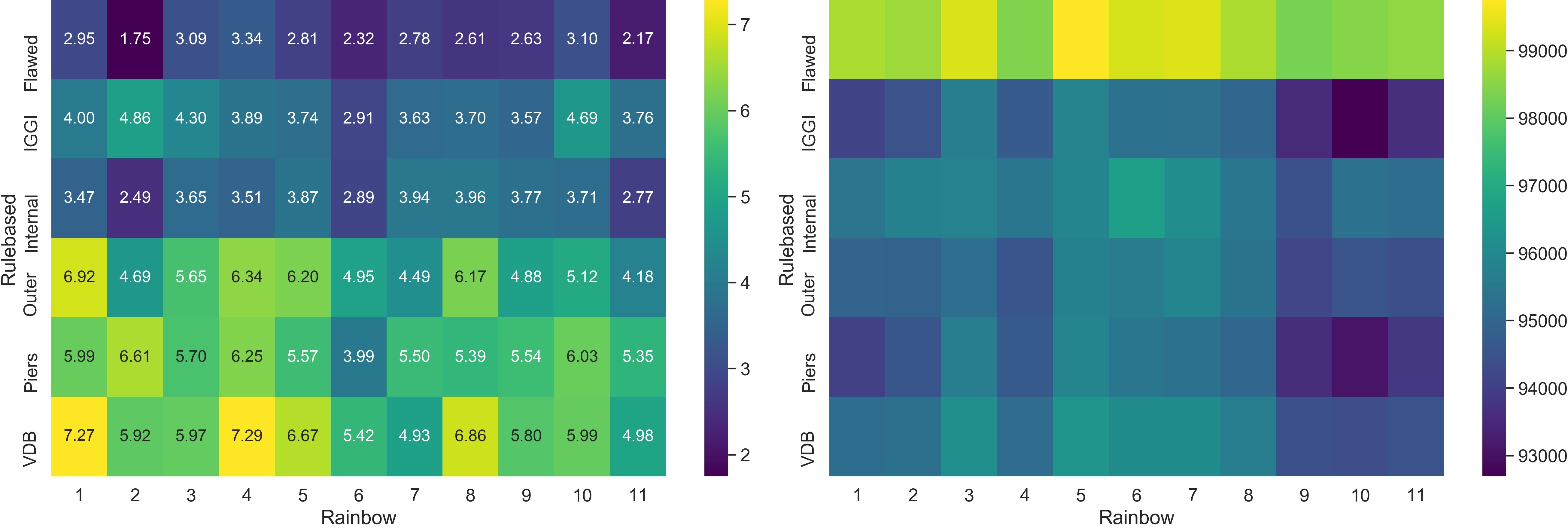}}
\label{fig3}
\caption{The left figure shows the average score of a two-player Hanabi game for 1,000 games. The vertical axes are six rule-based agents, and the horizontal axes are the Rainbow agents. The right figure show the behavioral differences between rule-based agents and Rainbow agents for 700 games.}
\end{figure*}

\begin{figure*}
\centerline{\includegraphics[width=0.95\textwidth]{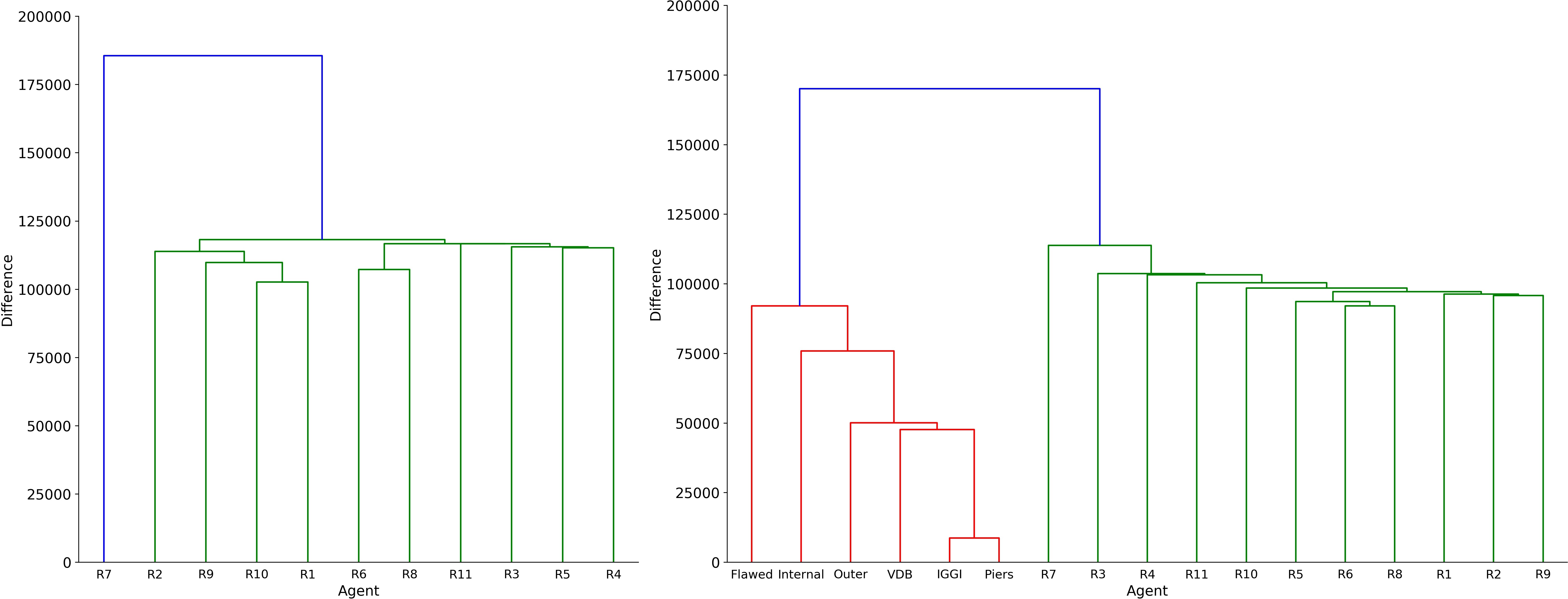}}
\label{fig4}
\caption{The left figure shows the dendrogram among the Rainbow agents. The right figure shows the dendrogram among the Rainbow and rule-based agents. 'R' indicates a 'Rainbow' agent and the following number is the index of Rainbow agents.}
\end{figure*}

\section{Experimental Results and Analysis}
\label{Experimental Results and Analysis}
\subsection{Hanabi Agents}
This section introduces the agents used in the experiments. There are two types of agents: reinforcement learning agents and rule-based agents. Rainbow agents are trained in Rainbow networks and are reinforcement learning agents. All other agents except are rule-based agents. $Internal$, $Outer$ agents were introduced in \cite{osawa2015solving}. Both agents remember the all information about the cards. The difference between $Internal$ and $Outer$ is former considers the player's own cards, otherwise the latter considers the other players' cards. $Van den Bergh$ agent was introduced in \cite{van2016aspects}. The $Van den Bergh$ considers the usefulness of whether a card can be one of the stacks or not. $IGGI$, $Piers$ agent were introduced in \cite{walton2017evaluating}. The characteristic of $IGGI$ is considering how to discard intelligently. The $Piers$ agent considers the additional condition rule to play more complex. Lastly, $Flawed$ agent was also introduced in \cite{walton2017evaluating}. The $Flawed$ agent has no strategic hint rules and risky action so that other player needs to give hint wisely. All the rule-based agents are set of specified rules. We describe the specified rules of rule-based agents in Table \ref{jeon.t2}. Additionally, we used an rule-based agents code implementation from github\footnote{https://github.com/rocanaan/hanabi-ad-hoc-learning}. 

\begin{itemize}
\item PlaySafeCard: Play a playable card.
\item OsawaDiscard: Discard a discard-able card.
\item TellPlayableCard: Tell another player about a playable card.
\item TellDispensable: Tell another player a card should not be discarded.
\item PlayProbablySafeCard(0.25): Play a card that is playable with probability 0.25.
\item DiscardOldestFirst: Discard the oldest card in your hand. 
\item TellPlayableCardOuter: Tell an other player an unknown hint about a playable card.
\item TellUnknown: Tell an other player any unknown hint about a card.
\item PlayIfCertain: Play a known card.
\item HailMary: If the deck and life tokens still remain, play a card that is playable with probability 0.0.
\item PlayProbablySafeCard(0.6): Play a card that is playable with probability 0.6.
\item TellAnyoneAboutUsefulCard: Tell another player about a useful unknown card.
\item DiscardProbablyUselessCard(0.99): Discard a card that is discard able with probability 0.99.
\item TellAnyoneAboutUselessCard: Tell another player about a useless unknown card.
\item TellMostInformation: Provide a hint that contains the most information.
\item DiscardProbablyUselessCard(0.0): Discard the most useless card with a threshold of 0.0.
\item TellRandomly: If an information token is available, tell a hint randomly.
\item DiscardRandomly: Discard a card randomly.
\end{itemize}

The Rainbow agent was created using six extensions of Deep Q-Networks\cite{mnih2015human}. Double DQN(DDQN)\cite{van2016deep} solves the Q-Networks' overestimation bias problem by separating the target and evaluation. Prioritized DDQN\cite{schaul2016prioritized} improved the sample efficiency by using state frequency to sample more important states. Dueling DDQN\cite{wang2016dueling} considered both state values and action advantages using two computations. Distributional DQN\cite{bellemare2017distributional} aimed to learn the distributions of returns, which are traditionally used to learn the expectation of returns. Lastly, Noisy DQN\cite{fortunato2018noisy} solves the limitations of high-dimensional action spaces by a adding noisy layer. Additionally, Rainbow network considers the multi-step target to bootstrap. Rainbow agents were seen to outperform all other agents.

\subsection{Rainbow Agent Ad-hoc Cooperation}
The Rainbow agents learned the two-player game for a total of 10,000 iterations with 10,000 steps for each iteration. The Rainbow network consisted of two fully-connected layers with 512 hidden units with 51 atoms to predict the value distributions. We trained via prioritized sampled data with a batch size of 32 and a discount factor of 0.99. We trained a total of 11 agents on each machine and proceeded learning two-player game mode. We considered the 9,950th and after of the total 10,000 iterations and selected them as experimental models. We implemented ad-hoc cooperation among 11 Rainbow agents. The results are reported in Figure 2. In Figure 2, most of the Rainbow agents play with unseen partners reasonably well, with the agents reported scores almost above 10. However, in the case of agent 7, despite its meaningful self-play score of 19.58, the agent failed to cooperate with the other agents. To analyze the Rainbow agents, we estimated the difference in max q-values among the Rainbow agents. The results show that agent 7 ultimately had a different decision process from the other agents, starting at about 50,000 states.

\subsection{Rule-based and Rainbow Agent Ad-hoc Cooperation}
The implemented rule-based agents are six, IGGI, Flawed, Piers, VDB, Internal, and Outer. We experimented with the these rule-based agents in combinations with Rainbow agents for 1,000 games as seen in Figure 3. In Figure 3, there are no Rainbow agents that achieved an ad-hoc performance better than 8. The highest score was 7.29 played by VDB and Rainbow 4. The results show only $Failure$ performances between the rule-based agents and the Rainbow agents. The right figure in Figure 3 shows the behavioral differences between the Rainbow and rule-based agents for 700 games and 52,000 states. Using \eqref{eq4}, we observed behavioral differences among the rule-based agents and Rainbow agents. The results show that differences occur above 45,000 states. The results of ad-hoc performance are given in Table \ref{jeon.t3}. All rule-based agents successfully cooperated with other rule-based agents, however, and Rainbow agents cannot cooperate with any other agents. To demonstrate the results for rule-based agents, we describe their self-play and ad-hoc performance in Table \ref{table4}. In Table \ref{table4}, the 'A1' and 'A2' columns represent the first and second agents, respectively, and the 'Ad-hoc' column is the ad-hoc cooperation score. Finally, the 'Self-play, A1' and 'Self-play, A2' columns are the self-play scores of the agents. The results of Table \ref{table4} show that rule-based agents exhibit successful performance, even with some ad-hoc performances close to or better than the self-play scores.

\begin{table}[!ht]
\renewcommand{\arraystretch}{1.3}
\caption{The Ad-hoc Performance Results}
\label{jeon.t3}
\centering
\begin{tabular}{|c|c|c|c|}
\hline
 Result & Rainbow & Rainbow+Rule-based & Rule-based\\ [0.3ex] 
\hline
 Failure & 55 & 33 & 0 \\ 
 Success & 0 & 0 & 15 \\ 
 Synergy & 0 & 0 & 0 \\ 
 \hline
\end{tabular}
\end{table}

\begin{table}[!ht]
\renewcommand{\arraystretch}{1.3}
\caption{The Analysis of Rule-based Agents Ad-hoc Performance}
\label{table4}
\centering
\begin{tabular}{|c|c|c|c|c|}
\hline
 A1 & A2 & Ad-hoc & Self-play, A1 & Self-play, A2\\ 
\hline
 Flawed & IGGI & 8.48 & 7.59 & 15.93 \\ 
 Flawed & Internal & 8.15 & 7.59 & 9.97 \\ 
 Flawed & Outer & 8.99 & 7.59 & 13.94 \\ 
 Flawed & Piers & 9.90 & 7.59 & 16.74 \\ 
 Flawed & VDB & 10.00 & 7.59 & 16.18 \\ 
 IGGI & Internal & 12.43 & 15.93 & 9.97 \\ 
 IGGI & Outer & 15.23 & 15.93 & 13.94 \\ 
 IGGI & Piers & 16.55 & 15.93 & 16.74 \\ 
 IGGI & VDB & 15.88 & 15.93 & 16.18 \\ 
 Internal & Outer & 11.85 & 9.97 & 13.94 \\ 
 Internal & Piers & 13.30 & 9.97 & 16.74 \\
 Internal & VDB & 13.21 & 9.97 & 16.18 \\
 Outer & Piers & 15.34 & 13.94 & 16.74 \\
 Outer & VDB & 15.53 & 13.94 & 16.18 \\
 Piers & VDB & 16.59 & 16.74 & 16.18 \\
 \hline
\end{tabular}
\end{table}

\begin{figure*}
\centerline{\includegraphics[width=0.9
\textwidth]{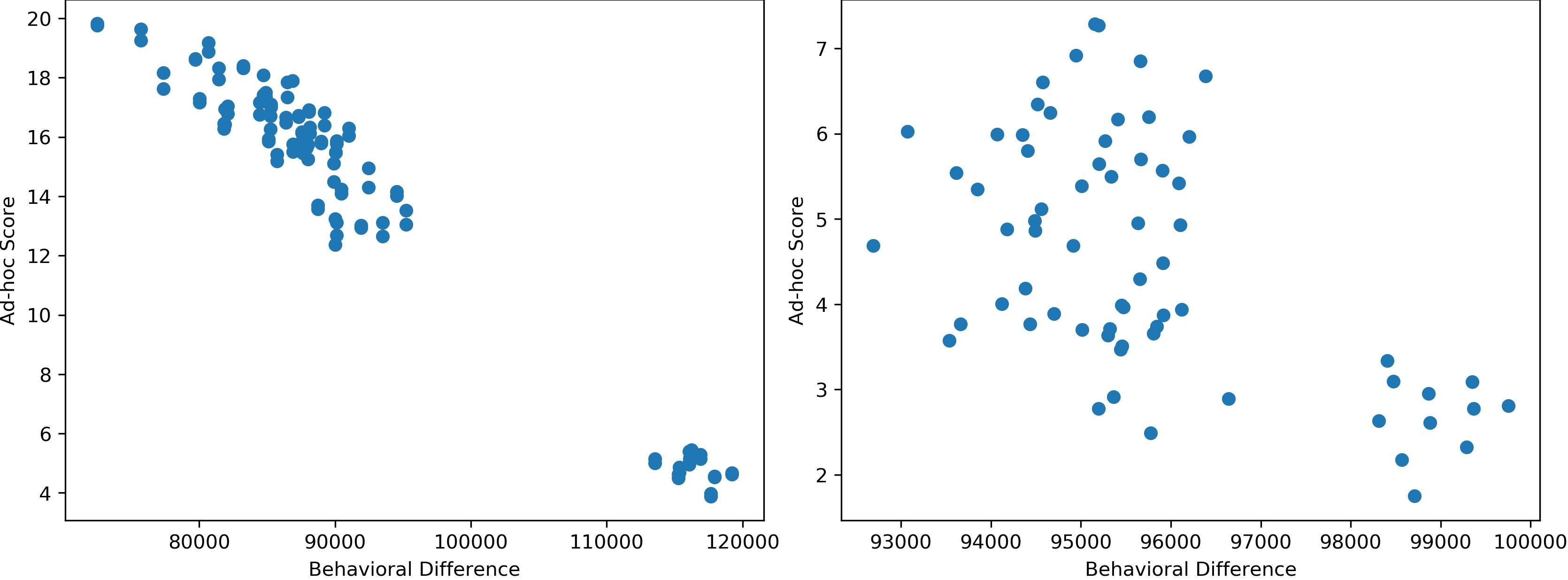}}
\label{fig6}
\caption{The figures show the Pearson correlation between behavioral difference and ad-hoc performance. The left figure shows the correlation of Rainbow agents. The right figure shows the correlation of Rainbow and rule-based agents.}
\end{figure*}

\begin{figure}[ht]
\centerline{\includegraphics[width=0.44
\textwidth]{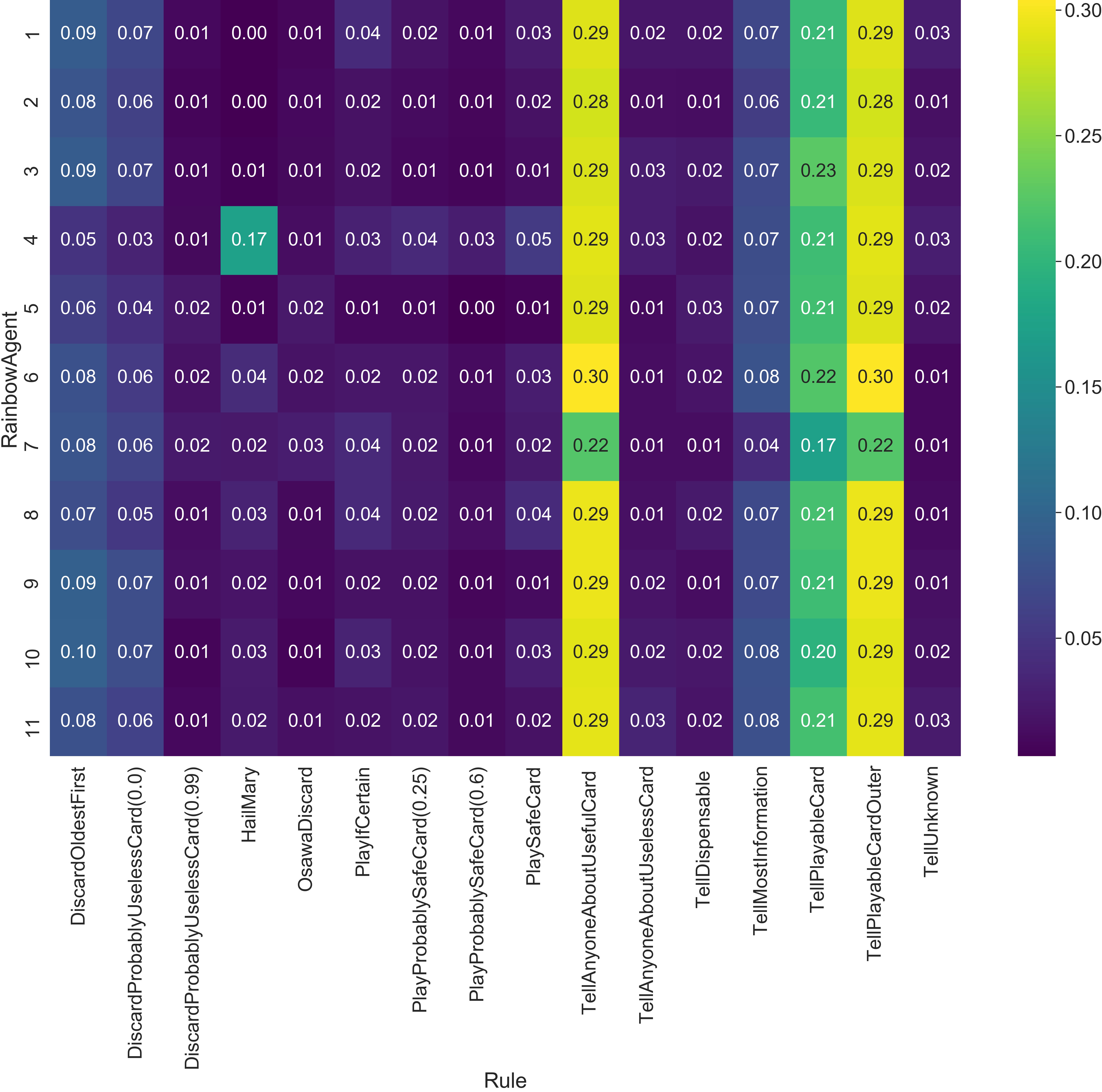}}
\label{fig5}
\caption{The figures show the similarity between action by the triggered rules and the Rainbow agents' actions for 1,000 states.}
\end{figure}

\section{Estimating Dissimilarity}
\label{Estimating Dissimilarity}

We measured the dissimilarity via hierarchical clustering to estimate the cluster distance of the Rainbow agents and the rule-based agents, as seen in Figure 4. The results show that Rainbow agent 7 results in a distinctly different cluster from the other Rainbow agents. Comparisons between the Rainbow and rule-based agents show that rule-based agents have some similarity with one another, which is due to the sub-rules the rule-based agents share. Despite the failure of Rainbow agents to cooperate with each other, the distance from the rule-based agents was too far, resulting in the Rainbow agents were grouped similarly.

We compare the dissimilarity between the triggered rules and Rainbow agents. This experiment aims to verify how similar the Rainbow agents are to the specified rules and ensure that there are no rules that interfere when cooperating with Rainbow agents. Except for the rules that make random action, we compared Rainbow agents 5 and 7 using 16 rules. We collected a thousand states for each rule when the specified rules were triggered. The experimental results show the triggered rules are quite different from the triggered rules' distributions. For a thousand states, there are no Rainbow agents who achieved above half similarity with any rules, as seen in Figure 6.

Lastly, we analyzed the Pearson correlations between behavioral difference and ad-hoc performance, as seen in Figure 5. In Figure 5, the correlation between Rainbow agents' behavioral differences and ad-hoc performance is -0.978, and its scatter plot is on the left-hand side of the figure, which indicates a strong relationship between behavioral difference and ad-hoc score. However, the correlation between Rainbow agents and rule-based agents is -0.554, which is much weaker than that of only Rainbow agents. We analyzed this reason for this and concluded it was caused by the ad-hoc scores of Rainbow and rule-based agents themselves being lower than those of the Rainbow agents, and thus the width of the score is relatively lower, as seen in the right-hand side of Figure 5. 

\section{Conclusion and Future Works}

In Hanabi, the ad-hoc cooperation problem occurs when agents learning via reinforcement learning cooperate with agents that were not seen in self-play settings. This problem is caused by behavioral differences. To demonstrate this, we first defined three outcomes for ad-hoc cooperation: $Failure$, $Success$, and $Synergy$. Subsequently, we analyzed the experimental results using hierarchical clustering, confirming that the agents were grouped into different groups due to behavioral differences. We also considered the Pearson correlation coefficients of agent behavioral differences and ad-hoc cooperation, demonstrating that the larger the behavioral difference, the lower the ad-hoc play performance. Additionally, we examined detailed rules for the rule-based agents and the behavioral differences of Rainbow agents to ensure that these specific rules did not negatively interfere with cooperation, thus confirming that the Rainbow agents differ in a fundamental way by utilizing different rules.

In future work, to overcome Rainbow agents' fixed actions regardless of the other agents considered, Diversity is All You Need(DIAYN)\cite{eysenbach2018diversity} may offer solutions to implementing diverse strategies. Since DIAYN can act in diverse ways in the context of skills, DIAYN may have a better attempt at handling diverse unseen agents. However, applying only DIAYN is not a solution, since the agent should be aware of the other agents' strategies and react accordingly. Therefore, DIAYN in conjunction with 'team aware' methods such as Bayesian Change Point Detection (BCPD)\cite{xuan2007bayesian} may be key to solving the ad-hoc cooperation problem in Hanabi. This result is not only limited in the card game Hanabi, but also provides clues to consider when playing with diverse agents, rather than just optimizing the performance of self-play in order to work with diverse agents in the multi-agent settings.


\bibliographystyle{IEEEtran}
\bibliography{references}

\end{document}